\pdfoutput=1

\documentclass[11pt]{article}

\usepackage[]{EMNLP2023}

\usepackage{times}
\usepackage{latexsym}

\usepackage[T1]{fontenc}

\usepackage[utf8
]{inputenc}

\usepackage{microtype}

\usepackage{inconsolata}

\usepackage{caption}
\usepackage{subcaption}
\usepackage{multirow}
\usepackage{booktabs}
\usepackage{makecell}
\usepackage{tablefootnote}

\usepackage[frozencache,cachedir=.]{minted}

\usepackage{amssymb}
\usepackage{amsmath}

\usepackage{pifont}
\newcommand{\cmark}{\ding{51}}
\newcommand{\xmark}{\ding{55}}

\usepackage[english, capitalise]{cleveref}

\usepackage{todonotes}

\presetkeys{todonotes}{inline}{}
\definecolor{very-light-gray}{gray}{0.97}
\presetkeys{todonotes}{size=footnotesize,backgroundcolor=very-light-gray}{}

\newcommand{\rparagraph}[1]{\vspace{1.2mm}\noindent\textbf{#1}}
\newcommand{\sparagraph}[1]{\vspace{0.0mm}\noindent\textbf{#1}}

\usepackage{xspace} 

\newcommand{\libname}{\textit{Adapters}\xspace}

\title{\textit{Adapters}: A Unified Library for\\ Parameter-Efficient and Modular Transfer Learning}

\author{
\bf Clifton Poth\thanks{\, Authors contributed equally.}$^{\, \, \, 1, 4}$, Hannah Sterz$^{*1}$, Indraneil Paul$^1$,  \\ 
{\bf Sukannya Purkayastha$^1$, Leon Engl\"ander$^1$, Timo Imhof$^1$, }  \\
{\bf  Ivan Vuli\'{c}$^{2}$, Sebastian Ruder$^{3}$, Iryna Gurevych$^{1}$, Jonas Pfeiffer$^{3}$} \\
$^1$Ubiquitous Knowledge Processing Lab,  Technical University of Darmstadt \\  
$^2$Language Technology Lab, University of Cambridge \hspace{0.5em} \\
$^3$Google DeepMind \hspace{0.5em} $^4$Cohere }

\begin{document}
\maketitle
\begin{abstract}

We introduce \textit{Adapters}, an open-source library that unifies parameter-efficient and modular transfer learning in large language models. By integrating 10 diverse adapter methods into a unified interface, \textit{Adapters} offers ease of use and flexible configuration. Our library allows researchers and practitioners to leverage adapter modularity through composition blocks, enabling the design of complex adapter setups. We demonstrate the library's efficacy by evaluating its performance against full fine-tuning on various NLP tasks. \libname provides a powerful tool for addressing the challenges of conventional fine-tuning paradigms and promoting more efficient and modular transfer learning. The library is available via \url{https://adapterhub.ml/adapters}.
\end{abstract}

\section{Introduction}
The ever-increasing size of pretrained large language models (LLMs) \citep{brownLanguageModelsAre2020a,chowdheryPaLMScalingLanguage2022} has made the established transfer learning paradigm of fine-tuning all model parameters on a downstream task \citep{howard-ruder-2018-universal, devlin-etal-2019-bert} extremely expensive.
Moreover, the requirement of \textit{parameter efficiency} at fine-tuning, while definitely paramount, is not the only shortcoming of the predominant LLM fine-tuning paradigm. It also suffers from other crucial issues such as negative interference, lack of positive transfer between tasks in multi-task learning~\citep{mccloskey1989catastrophicinterference}, catastrophic forgetting \citep{french1999catastrophic}, and poor generalization.

Two closely related lines of research aimed at addressing this set of challenges have gained significant attention recently. First, \textit{parameter-efficient fine-tuning} \citep{lialinScalingScaleGuide2023,Sabry2023PEFTRefAM} focuses on the aspect of computational efficiency and feasibility by only fine-tuning a small number of parameters for downstream tasks.
Second, \textit{modular transfer learning} \citep{pfeiffer2023modulardeeplearning} focuses on the aspect of knowledge transfer by designing self-contained network modules which can be aggregated for better multi-task performance and generalization.
In practice, these often represent two sides of the same coin.
Methods that devise small components within a language model for fine-tuning on labeled task data, henceforth generally denoted as \textit{adapters}, are both parameter-efficient \textit{and} modular in nature.

The initial release of \textit{AdapterHub} \citep{pfeiffer-etal-2020-adapterhub} marks the first attempt to systematically make adapters accessible to researchers and practitioners in an easy-to-use framework.
AdapterHub proposed a framework to easily integrate, train and use adapters for state-of-the-art Transformer models with minimal changes.
It additionally established an open platform to share, explore and consume pre-trained adapter modules.
While AdapterHub focused on bottleneck-style adapters \citep{houlsbyParameterefficientTransferLearning2019} initially, the field of adapter methods has expanded substantially since \citep[\textit{among others}]{li-liang-2021-prefix,mahabadiCompacterEfficientLowRank2021a,huLoRALowRankAdaptation2022,heUnifiedViewParameterEfficient2022,liuFewShotParameterEfficientFineTuning2022}.

With increasing interest in adapter methods, new tools and libraries have been developed. \textit{OpenDelta} \citep{hu-etal-2023-opendelta}, HuggingFace's \textit{PEFT} \citep{peft} and \textit{LLM-Adapters} \citep{Hu2023LLMAdaptersAA} are recent examples of libraries which attempt to unify adapter methods in a single code base and extend their applicability to new model architectures.
However, these works exclusively focus on the parameter-efficiency aspect of adapters, neglecting the modularity side of these methods.

\rparagraph{Contributions.}
Based on the initial version of AdapterHub, we, therefore, propose \textit{Adapters}, a new library aimed at \textit{unifying parameter-efficient and modular transfer learning}.
Compared to the first AdapterHub iteration and concurrent libraries, our main contributions can be summarized as follows:
\textbf{1)}~We propose a self-contained library that integrates 10 diverse adapter methods into a unified interface for easy usage and flexible configuration;
\textbf{2)}~we develop a simple way of leveraging the modularity of adapters by designing \textit{composition blocks} that allow flexibly defining complex adapter setups;
\textbf{3)}~we integrate all methods into 20 Transformer-based models spanning NLP, vision, and multi-modal applications;
\textbf{4)}~we evaluate the performance of our adapter implementations against full fine-tuning on a diverse set of tasks.

\section{Background}

We use the term \textit{adapter} in a more general sense to refer to a broad family of transfer learning methods that share the two defining properties: parameter efficiency and modularity. For a detailed overview of different adapter architectures, we refer the reader to the recent survey by \citet{pfeiffer2023modulardeeplearning}.

\subsection{Parameter Efficiency}

Let the parameters of a language model be composed of a set of pre-trained parameters $\Theta$ (frozen) and a set of parameters $\Phi$ (where $\Phi$ can either be newly introduced or $\Phi \subset \Theta$).
During fine-tuning, adapter methods optimize only $\Phi$ according to a loss function $L$ on a dataset $D$:
$$
\Phi^* \leftarrow \arg \min_{\Phi} L(D; \{\Theta, \Phi\})
$$
\noindent Different adapter methods insert parameters $\Phi$ at different locations of a pre-trained large model.
Bottleneck adapters \citep{Rebuffi2017LearningMV,houlsbyParameterefficientTransferLearning2019}, as one of the early methods, introduce bottleneck feed-forward layers in each layer of a Transformer model.
Subsequent designs have adapted a Transformer model's self-attentions \citep{li-liang-2021-prefix}, bias terms \citep{ben-zaken-etal-2022-bitfit}, input prompts \citep{lester-etal-2021-power} or embeddings \citep{pfeiffer-etal-2021-unks}.
Complementary lines of work have focused on optimizing the parameter efficiency \citep{mahabadiCompacterEfficientLowRank2021a,liuFewShotParameterEfficientFineTuning2022} and runtime efficiency \citep{huLoRALowRankAdaptation2022,Lei2023ConditionalAdapters} of adapters or have attempted to unify multiple components in a single framework \citep{heUnifiedViewParameterEfficient2022,mao-etal-2022-unipelt}.

\subsection{Modularity}

A modular deep learning model is composed of modules that each capture a specific functionality of the full model, such as task or language capacities.
\citet{pfeiffer2023modulardeeplearning} propose a taxonomy of modular deep learning methods covering the dimensions of computation function, routing, aggregation, and training.

Routing and aggregation are of special interest here as they coordinate the composition of multiple adapter modules, a key functionality enabled by modularity.
Exemplary existing work includes using stochastic routing through adapters \citep{wang-etal-2022-adamix}, adapter parameter averaging \citep{friedman-etal-2021-single}, sequential function aggregation of adapter modules \citep{pfeiffer-etal-2022-lifting} as well as weighted \citep{wang-etal-2021-efficient-test} and attention-based \citep{pfeiffer-etal-2021-adapterfusion} output aggregation.

Finally, along the training dimension, the modularity of adapters allows using pre-trained adapter modules as initialization for further fine-tuning \citep{poth-etal-2021-pre,vu-etal-2022-spot}.

\section{The \libname Library}

\begin{table}[!t]
    \centering
    \def\arraystretch{0.85}
    \resizebox{\linewidth}{!}{
    \begin{tabular}{ccc}
        \toprule
        & \textbf{AdapterHub v1} & \textbf{Adapters} \\
        \midrule
        Design & \makecell{Fork of\\ Transformers} & \makecell{Self-contained\\ add-on library} \\
        \makecell{Adapter methods} & 2 & 10 \\
        \makecell{Complex\\ configurations} & \xmark & \cmark \\
        \makecell{Composition blocks} & \xmark \tablefootnote{V1 already supported stacking and fusing adapter, however without flexibly composable blocks.} & \cmark \, (6) \\
        \makecell{Model architectures} & 3 & 20 \\
        \makecell{AdapterHub.ml /\\ HF Hub integration} & \cmark \, / \, \xmark & \cmark \, / \, \cmark \\
        \bottomrule
    \end{tabular}
    }
    \vspace{-1mm}
    \caption{Feature comparison between the initial \textit{AdapterHub} release \citep{pfeiffer-etal-2020-adapterhub} and the proposed \libname library.}
    \label{tab:feature_table}
    \vspace{-3.5mm}
\end{table}

\libname builds on many design decisions established in the initial \textit{AdapterHub} release \citep{pfeiffer-etal-2020-adapterhub}, but offers substantial extensions both `horizontally' (e.g., extending the support to many more pretrained neural architectures, extending the coverage of adapter architectures) and `vertically' (e.g., adding new composition and processing capabilities).
\cref{tab:feature_table} gives an overview of the differences between the initial \textit{AdapterHub} and \libname.
The core features adopted from the initial release, facilitating its ease of use and wider adoption by researchers and practitioners, include: \textbf{1)} Tight integration into the widely used HuggingFace Transformers \citep{wolf-etal-2020-transformers} library; \textbf{2)} adaptation of pre-existing Transformers fine-tuning scripts with minimal changes; \textbf{3)} single-line saving and loading of adapter modules from a shared community hub.

\begin{listing}[!t]
\begin{center}
\begin{minted}[fontsize=\small]{python}
import adapters
from transformers import AutoModel

model = AutoModel.from_pretrained("..")
adapters.init(model)
model.add_adapter("a", config="seq_bn")
model.add_adapter("b", config="seq_bn")
model.train_adapter(Parallel("a", "b"))
\end{minted}
\end{center}
\caption{Example of adding adapters to an existing Transformers model. After model instantiation, \texttt{init()} introduces adapter-specific functionality. Two bottleneck adapters are added via \texttt{add\_adapter()} and activated for parallel training.}
\label{listing:code_example}
\vspace{-3.5mm}
\end{listing}

\subsection{Transformers Integration}

Unlike the initial AdapterHub, \libname is designed as a standalone package that acts as an add-on to the Transformers library.
\libname provides adapter implementations and management methods that can be injected into pre-trained Transformer checkpoints without modifying the original model code directly.
We provide two approaches to this end: (i) by attaching to existing models and (ii) by providing our own, specialized model classes.

\rparagraph{Attaching to Models.}
The \verb|init()| method provides a straightforward solution for making adapters accessible to pre-existing model classes post-hoc.
A model checkpoint for one of the supported architectures (cf. §~\ref{sec:supported_models}) can be instantiated via any of the model classes provided by the Transformers library.
An example of this approach is given in \cref{listing:code_example}.
All adapter-related functionality is injected post-instantiation by passing the model instance to the \verb|init()| method.
Afterwards, all methods provided by \libname are easily invokable from the same model instance.

\rparagraph{AdapterModel Classes.}
As an alternative to post-hoc initialization, \libname provides a set of built-in model classes optimized for working with adapters.
Following HuggingFace's naming conventions, these classes are named \verb|XXXAdapterModel|, where \verb|XXX| is the name of the model architecture.
Compared to models with attached adapters, these classes especially provide more flexibility with regard to prediction heads.
Each model instance can have multiple named prediction heads targeted towards different tasks loaded simultaneously.
These prediction heads can be associated with adapter modules by sharing a common name.

Providing this functionality is crucial for enabling features such as quickly switching between adapters targeted toward different tasks at runtime.
It is also essential for creating composed adapter setups such as parallel inference on multiple tasks (cf. §~\ref{sec:adapter_composition}).
We, therefore, provide automatic conversion from HuggingFace's model classes - typically paired with a single, fixed prediction head - to our newly introduced classes featuring flexible prediction heads.

\subsection{Unified Adapter Interface}

\libname defines a common interface of methods covering the full life cycle of working with adapters.
This includes methods for adding, activating, saving, releasing, loading, aggregating, and deleting adapter modules.
When adding a new adapter to a model (e.g., via \verb|add_adapter()|), it is given a unique identifier string.
All adapter-related methods then solely use this string to identify the adapter module an operation should be performed on.
Thus, the adapter interface at the model level can be agnostic to specific adapter methods.

This interface, as well as adapter implementations at the module level, are integrated into model classes via Python mixins and dynamically modifying Python classes at runtime to keep changes to the existing Transformers code base minimal.

\begin{figure*}[ht!]
    \centering
    \begin{subfigure}[b]{0.11\textwidth}
        \centering
        \includegraphics[width=\textwidth]{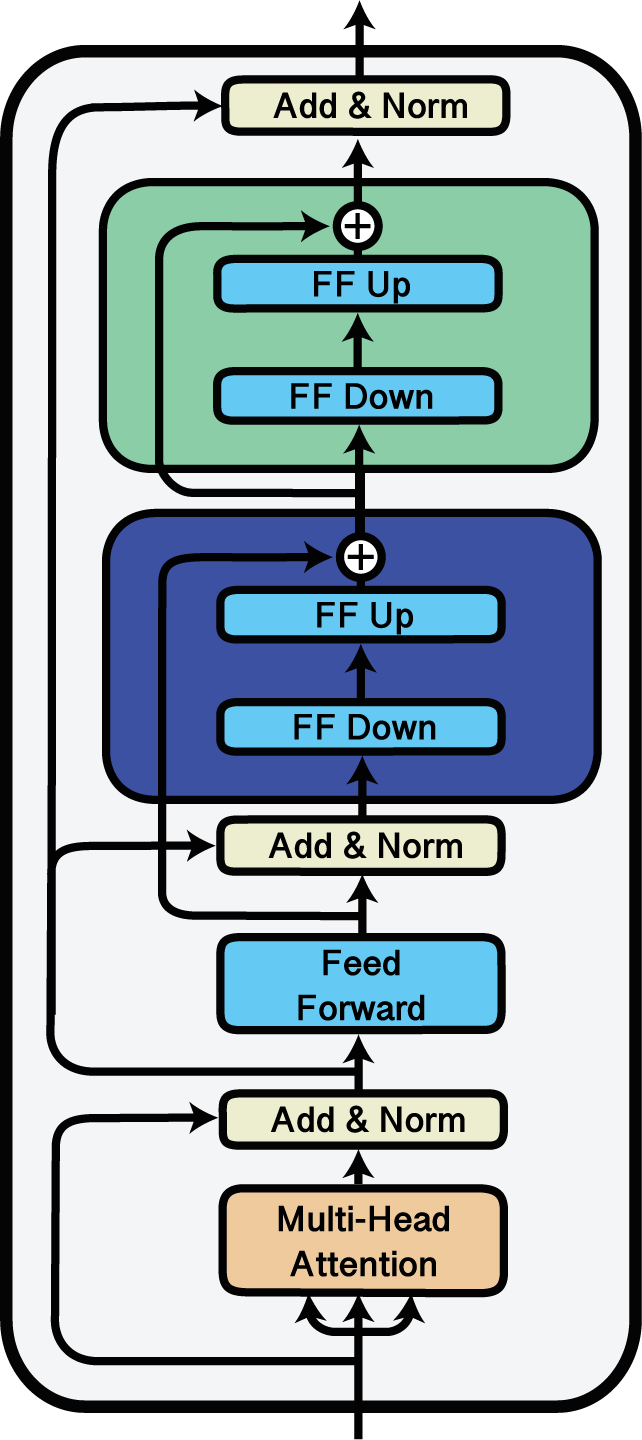}
        \caption{Stack}
    \end{subfigure}
    \hfill
    \begin{subfigure}[b]{0.17\textwidth}
        \centering
        \includegraphics[width=\textwidth]{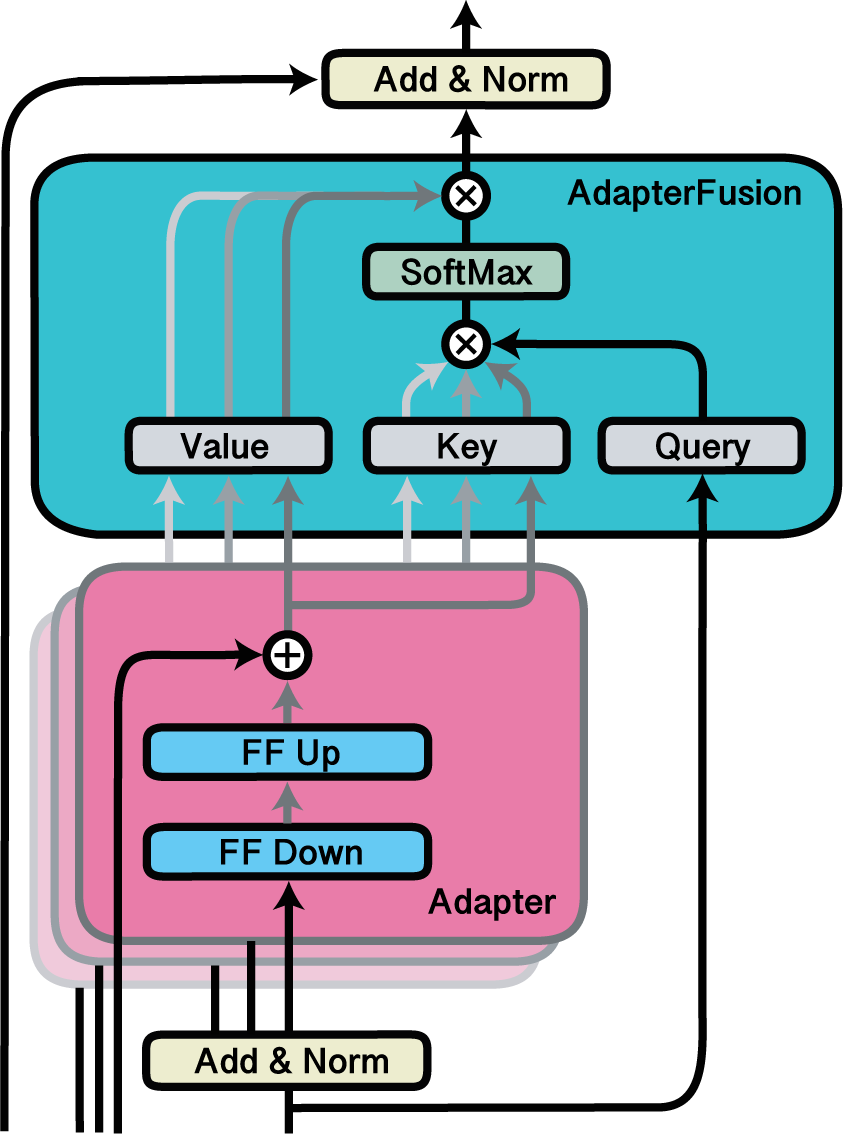}
        \caption{Fuse}
    \end{subfigure}
    \hfill
    \begin{subfigure}[b]{0.22\textwidth}
        \centering
        \includegraphics[width=\textwidth]{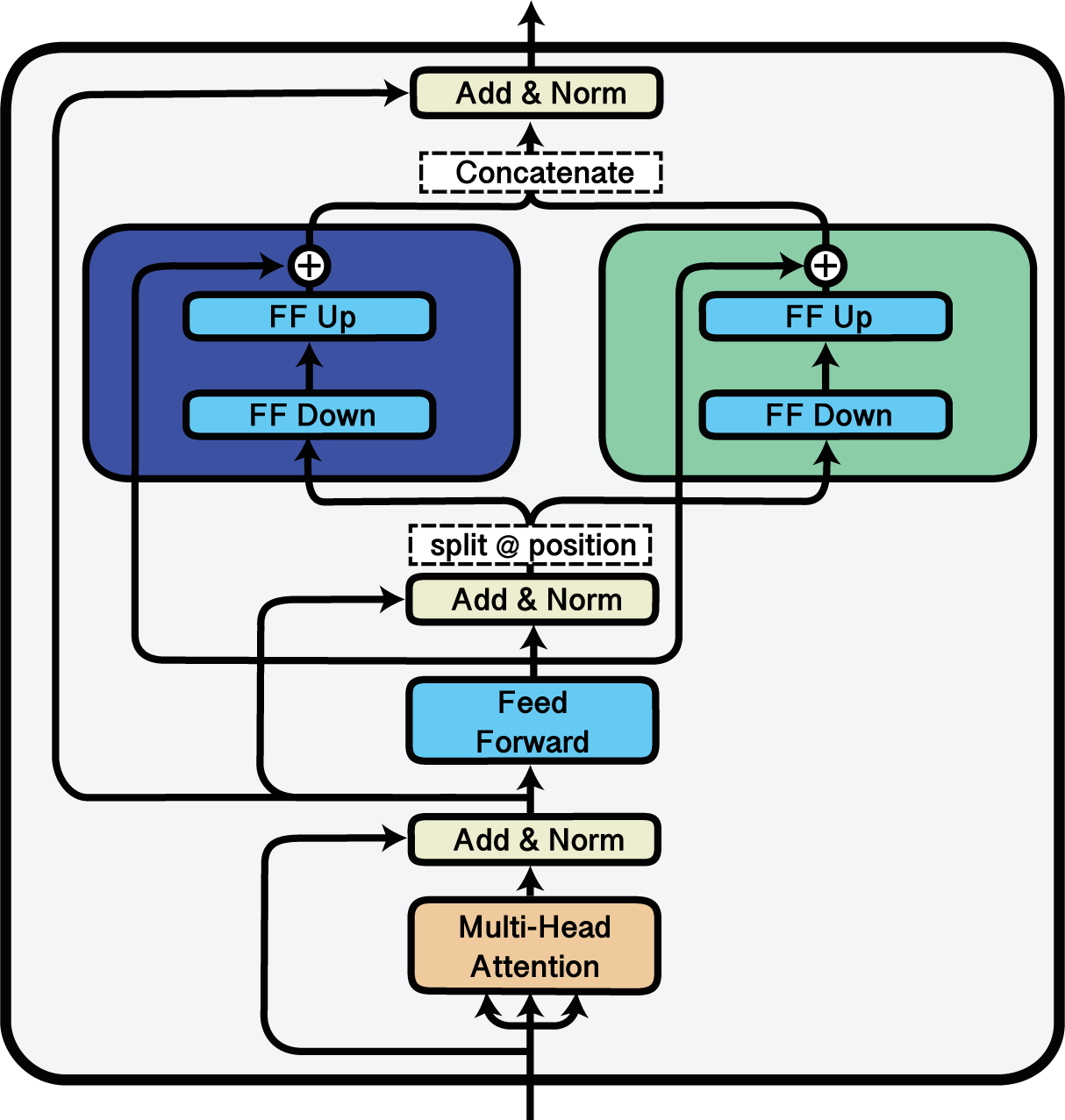}
        \caption{Split}
    \end{subfigure}
    \hfill
    \begin{subfigure}[b]{0.145\textwidth}
        \centering
        \includegraphics[width=\textwidth]{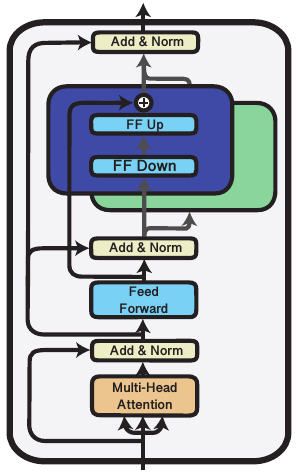}
        \caption{BatchSplit}
    \end{subfigure}
    \hfill
    \begin{subfigure}[b]{0.105\textwidth}
        \centering
        \includegraphics[width=\textwidth]{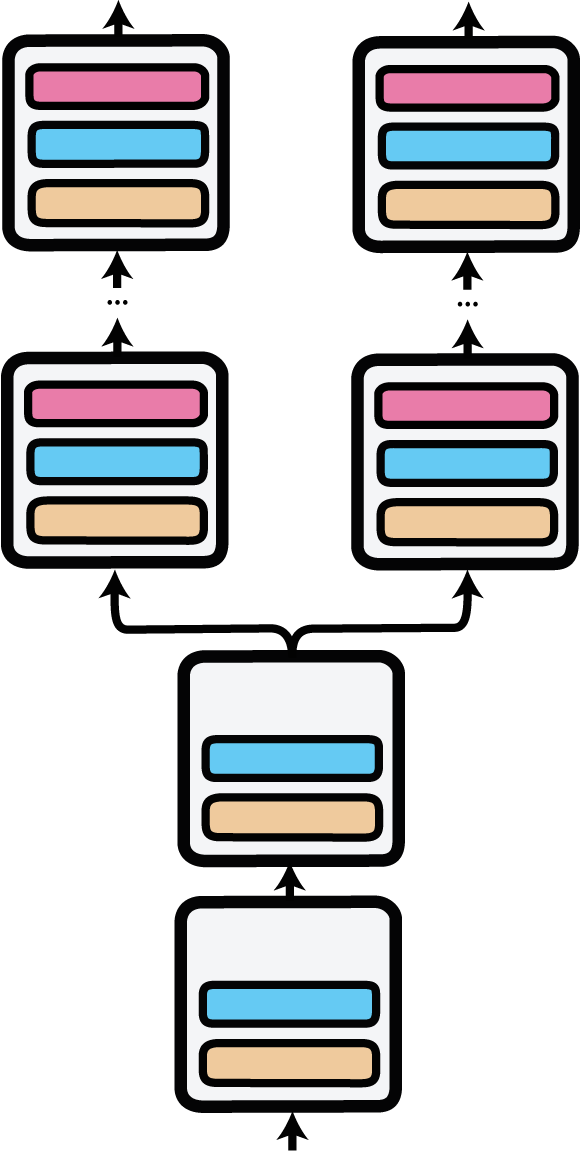}
        \caption{Parallel}
    \end{subfigure}
    \hfill
    \begin{subfigure}[b]{0.22\textwidth}
        \centering
        \includegraphics[width=\textwidth]{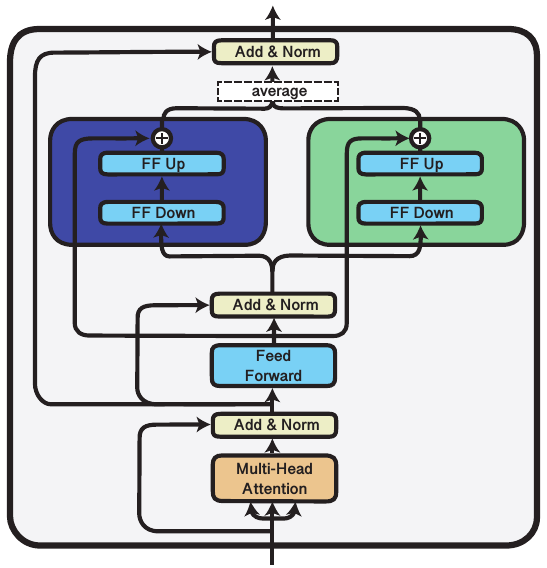}
        \caption{Average}
    \end{subfigure}
    \caption{Overview of adapter composition blocks supported in the \libname library at the time of writing this paper.}
    \label{fig:composition_blocks}
    \vspace{-2.5mm}
\end{figure*}

\subsection{Adapter Methods}

\begin{table}[t!]
    \centering
    \def\arraystretch{0.9}
    \resizebox{\linewidth}{!}{
    \begin{tabular}{lll}
    \toprule
    \textbf{Method name} & \textbf{Default config}\\
    \midrule
    Bottleneck adapter \citep{houlsbyParameterefficientTransferLearning2019} & \texttt{[double\_]seq\_bn} \\
    Invertible adapter \citep{pfeiffer-etal-2020-mad} & \texttt{seq\_bn\_inv} \\
    Prompt tuning \citep{lester-etal-2021-power} & \texttt{prompt\_tuning} \\
    Prefix tuning \citep{li-liang-2021-prefix} & \texttt{prefix\_tuning} \\
    Compacter \citep{mahabadiCompacterEfficientLowRank2021a} & \texttt{compacter} \\
    LoRA \citep{huLoRALowRankAdaptation2022} & \texttt{lora} \\
    (IA)³ \citep{liuFewShotParameterEfficientFineTuning2022} & \texttt{ia3} \\
    Parallel adapter \citep{heUnifiedViewParameterEfficient2022} & \texttt{par\_bn} \\
    Mix-and-Match adapter \citep{heUnifiedViewParameterEfficient2022} & \texttt{mam} \\
    UniPELT \citep{mao-etal-2022-unipelt} & \texttt{unipelt} \\
    \bottomrule
    \end{tabular}
    }
    \caption{Overview of adapter methods supported in the \libname library at the time of submission (Aug 2023).}
    \label{tab:methods}
    \vspace{-2.5mm}
\end{table}

Each adapter method is defined by a configuration object or string which allow flexible customization of various properties of an adapter module, including placement, capacity, residual connections, initialization etc.
We distinguish between single methods consisting of one type of adapter module and complex methods consisting of multiple different adapter module types.
\Cref{tab:methods} gives an overview of all methods currently integrated into \libname, along with their configuration strings.

\subsubsection{Single Methods}
\label{subsec:simple_peft}

\libname supports single adapter methods that introduce parameters in new feed-forward modules such as bottleneck adapters \citep{houlsbyParameterefficientTransferLearning2019}, introduce prompts at different locations such as prefix tuning \citep{li-liang-2021-prefix}, reparameterize existing modules such as LoRA \citep{huLoRALowRankAdaptation2022} or re-scale their output representations such as (IA)³ \citep{liuFewShotParameterEfficientFineTuning2022}.
Detailed descriptions for all currently implemented single methods are given in \cref{sec:adapter_descriptions}, see also \Cref{tab:methods} again.

\subsubsection{Complex Methods}

While different efficient fine-tuning methods and configurations have often been proposed as standalone, combining them for joint training has proven to be beneficial \cite{heUnifiedViewParameterEfficient2022, mao-etal-2022-unipelt}. To make this process easier, \libname provides the possibility to group multiple configuration instances using the \verb|ConfigUnion| class.
This flexible mechanism allows easy integration of multiple complex methods proposed in the literature (as the two examples outlined below), as well as the construction of other, new complex configurations currently not available nor benchmarked in the literature~\cite{autopeft}.




\rparagraph{Mix-and-Match Adapters \citep{heUnifiedViewParameterEfficient2022}} was proposed as a combination of Prefix-Tuning and parallel bottleneck adapters.
Using \verb|ConfigUnion|, this method can be defined as:
\vspace{-1.5mm}
\begin{minted}[fontsize=\small]{python}
config = ConfigUnion(
  PrefixTuningConfig(bottleneck_size=800),
  ParallelConfig(),
)
model.add_adapter("name", config=config)
\end{minted}

\rparagraph{UniPELT \citep{mao-etal-2022-unipelt}} combines LoRA, Prefix Tuning, and bottleneck adapters in a single unified setup.
It additionally introduces a gating mechanism that controls the activation of the different adapter modules.
$\mathcal{G}_m \leftarrow \sigma(W_{\mathcal{G}_m} \cdot x)$.

\subsection{Adapter Composition}\label{sec:adapter_composition}

While the modularity and composability aspects of adapters have seen increasing interest in research, existing open-source libraries \citep{peft, hu-etal-2023-opendelta} have largely overlooked these aspects.
\libname makes adapter compositions a central and accessible part of working with adapters by enabling the definition of complex, composed adapter setups.
We define a set of simple \textit{composition blocks} that each capture a specific method of aggregating the functionality of multiple adapters.
Each composition block class takes a sequence of adapter identifiers plus optional configuration as arguments.
The defined adapter setup is then parsed at runtime by \libname to allow for dynamic switching between adapters per forward pass. \cref{fig:composition_blocks} shows schematic illustrations of all composition blocks supported by \libname. 
\cref{listing:composition_blocks} shows examples of how the same composition blocks are defined in code:

\begin{listing}[!ht]
\begin{minted}[fontsize=\small]{python}
Stack("a", "b", "c")
Fuse("d", "e", "f")
Split("g", "h", splits=[64, 64])
BatchSplit("i", "j", batch_sizes=[2, 4])
Parallel("k", "l", "m")
Average("n", "o", weights=[0.3, 0.7])
Stack("p", Parallel("q", "r"))
\end{minted}
\vspace{-4mm}
\caption{Code examples of composition blocks supported by \libname. Strings represent adapter IDs.}
\label{listing:composition_blocks}
\vspace{-3mm}
\end{listing}
\noindent In what follows, we present each supported composition in more detail.

\rparagraph{Stack.}
The Stack block allows stacking multiple adapters sequentially within a Transformer layer.
This type of composition is, e.g., used in the MAD-X framework for cross-lingual transfer \citep{pfeiffer-etal-2020-mad}, where language and task adapters are stacked.
In \cref{listing:composition_blocks}, the input is first passed through \textit{a}, the output of \textit{a} is then inputted to \textit{b}, and the output of \textit{b} is finally inputted to \textit{c}.

\rparagraph{Fuse.}
The Fuse block can be used to activate an AdapterFusion layer \citep{pfeiffer-etal-2021-adapterfusion}.
AdapterFusion is a non-destructive way to combine the knowledge of multiple pre-trained adapters on a new downstream task.
In \cref{listing:composition_blocks}, we activate the adapters \textit{d}, \textit{e}, and \textit{f} as well as the fusion layer that combines the outputs of all three.\footnote{Note that this requires a fusion layer to be added beforehand via \texttt{add\_adapter\_fusion()}.}

\rparagraph{Split.}
The Split block can be used to split an input sequence between multiple adapters.
This e.g. enables splitting multimodal input sequences to modality-specific adapters \citep{pfeiffer-etal-2022-xgqa}.
In \cref{listing:composition_blocks}, we split each input sequence between adapters \textit{g} and \textit{h}.
All tokens with indices 0 - 63 are forwarded through \textit{g} while the next 64 tokens beginning at index 64 are forwarded through \textit{h}.

\rparagraph{BatchSplit.}
The BatchSplit block splits inputs along the batch size dimension between several adapters.
That is, different adapters receive different sub-batches of the full input batch.
In \cref{listing:composition_blocks}, we split the input batch between adapters \textit{i}, and \textit{j}.
Adapter \textit{i} receives two sequences and \textit{j} receives four sequences.
The sum of all specified sub-batches has to match the batch size of the input.

\rparagraph{Parallel.}
This block can be used to enable independent parallel training and inference on different adapters, where each adapter has its own prediction head.
The implementation automatically replicates all inputs at the first occurrence of parallel adapter modules, sharing the inputs in all lower layers without parallel modules.
This mechanism was first used in \citet{ruckle-etal-2021-adapterdrop}.
In \cref{listing:composition_blocks}, we forward all inputs via adapters \textit{k}, \textit{l}, and \textit{m} in parallel.

\rparagraph{Average.}
Following approaches of ensembling full models at inference time for better generalization, recent work has explored methods of averaging pre-trained adapters.
This includes averaging adapter output representations \citep{wang-etal-2021-efficient-test} as well as averaging adapter parameters \citep{friedman-etal-2021-single,wang-etal-2022-adamix,chronopoulou-etal-2023-adaptersoup}.
\libname provides built-in support for both types of inference time-averaging methods.
\textbf{Output averaging} allows to dynamically aggregate the output representations of multiple adapters in a model forward pass via weighted averaging.
This is realized via the Average composition block.
In \cref{listing:composition_blocks}, two adapters are averaged with the weights $0.3$ for \textit{n} and $0.7$ for \textit{o}.
\textbf{Parameter averaging} enables creating a new adapter via weighted averaging of the parameters of multiple pre-trained adapters.
As this process is typically not done dynamically at runtime, \libname provides \verb|average_adapter()| as a dedicated method.
Compared to output averaging, parameter averaging of adapters has the advantage of not inducing any additional inference time relative to using a single adapter.

\rparagraph{Nesting.}
Finally, it is possible to nest composition blocks within other composition blocks to create deeper and more complex compositions.
\libname defines a set of allowed nestings to restrict the users to setups that are sensible.
As an example, we nest a Parallel block within a Stack block in \cref{listing:composition_blocks}.

\subsection{Supported Models}\label{sec:supported_models}

At the time of release, \libname has built-in support for 20 widely adopted model architectures included in the Transformers library.
This covers text encoder models such as BERT \citep{devlin-etal-2019-bert} and DeBERTa \citep{heDebertaDecodingEnhancedBert2021}, text decoder models such as GPT-2 \citep{Radford2019LanguageMA}, sequence-to-sequence models such as BART \citep{lewis-etal-2020-bart} and T5 \citep{raffelExploringLimitsTransfer2020a}, vision encoder models such as ViT \citep{dosovitskiyImageWorth16x162021a}, as well as multimodal models such as CLIP \citep{radfordLearningTransferableVisual2021}.\footnote{An up-to-date full list of supported models can be found at \url{https://docs.adapterhub.ml/model_overview.html}.}

While adapter-related implementations mostly can be shared across all supported models, correctly integrating them into each model implementation requires manual effort.
While it is difficult to standardize this process due to differences between model architectures, we provide clear guidelines for integrating new models in the form of shared interfaces and step-by-step documentation\footnote{\url{https://docs.adapterhub.ml/contributing/adding_adapters_to_a_model.html}}.

\begin{table*}[t!]
    \centering
    \def\arraystretch{0.87}
    \resizebox{\linewidth}{!}{
    \begin{tabular}{rcccccccccccc}

\toprule
\multirow{3}{*}{\textbf{Method}} & \multicolumn{7}{c}{\textbf{Sequence Classification}} & \textbf{Regression} & \textbf{Mult. Choice} &      \textbf{Extract. QA} & \textbf{Tagging} & \multirow{3}{*}{\textbf{Avg.}} \\
\cmidrule(l){2-8} \cmidrule(l){9-9} \cmidrule(l){10-10} \cmidrule(l){11-11} \cmidrule(l){12-12} 
 &  \textbf{CoLA} &  \textbf{MNLI} &  \textbf{MRPC} &  \textbf{QNLI} &   \textbf{QQP} &   \textbf{RTE} &  \textbf{SST-2} &  \textbf{STS-B} & \textbf{Cosmos QA} & \textbf{SQuAD v2} &  \textbf{CoNLL-2003} & \\

& \texttt{Dev MCC} & \texttt{Dev Acc.} & \texttt{Dev F1} & \texttt{Dev Acc.} & \texttt{Dev F1} & \texttt{Dev Acc.} & \texttt{Dev Acc.} & \texttt{Dev PCC} & \texttt{Dev Acc.} & \texttt{Dev F1} & \texttt{Test F1} & \\
 
\midrule

\multirow{2}{*}{\texttt{double\_seq\_bn}} & 63.58 & 87.61 & \textbf{93.31} & 92.84  & 91.58  & \textbf{80.87} & 94.73 &    90.85        & 70.99   &  \textbf{84.89}   & 91.92 & \textbf{85.74} \\
 & \small{($\pm$19.19)} & \small{($\pm$26.41)} & \small{($\pm$4.52)} & \small{($\pm$17.17)} & \small{($\pm$36.83)} & \small{($\pm$11.09)} & \small{($\pm$17.51)} & \small{($\pm$27.16)} & \small{($\pm$16.87)} & \small{($\pm$5.52)} & \small{($\pm$17.65)} & \small{($\pm$18.17)} \\
\cmidrule(lr){2-13}

\multirow{2}{*}{\texttt{seq\_bn}} & \textbf{71.22} & 87.5 & 92.91 & \textbf{93.15} & 89.69 & 79.42 & 95.18 & 89.44 & 69.68         & 82.88  &  \textbf{92.02} & \textbf{85.74} \\
& \small{($\pm$23.40)} & \small{($\pm$20.39)} & \small{($\pm$4.54)} & \small{($\pm$15.83)} & \small{($\pm$21.31)} & \small{($\pm$9.81)} & \small{($\pm$13.26)} & \small{($\pm$20.33)} & \small{($\pm$16.44)} & \small{($\pm$\textbf{1.04})} & \small{($\pm$11.48)} & \small{($\pm$14.34)} \\
\cmidrule(lr){2-13}

\multirow{2}{*}{\texttt{par\_bn}} & 63.95 & 87.44 & 93.24 & 93.04 & 88.32 & 77.98 & 94.95 & 90.33 & \textbf{80.10}          & 82.56    &    91.95   & 85.81  \\
& \small{($\pm$23.72)} & \small{($\pm$21.66)} & \small{($\pm$4.65)} & \small{($\pm$17.26)} & \small{($\pm$33.14)} & \small{($\pm$10.95)} & \small{($\pm$16.97)} & \small{($\pm$\textbf{5.64})} & \small{($\pm$18.47)} & \small{($\pm$6.70)} & \small{($\pm$27.60)} & \small{($\pm$16.98)}  \\
\cmidrule(lr){2-13}

\multirow{2}{*}{\texttt{compacter}} & 55.52 & 86.10 & 90.43 & 92.42 & 86.68 & 68.59 & 94.15 & 90.06 &           67.91 &    79.20 &     91.27 & 82.03 \\
& \small{($\pm$\textbf{13.67})} & \small{($\pm$\textbf{1.99})} & \small{($\pm$\textbf{3.58})} & \small{($\pm$\textbf{2.68})} & \small{($\pm$\textbf{2.14})} & \small{($\pm$\textbf{4.91})} & \small{($\pm$\textbf{0.81})} & \small{($\pm$23.27)} & \small{($\pm$10.42)} & \small{($\pm$8.87)} & \small{($\pm$8.58)} & \small{($\pm$\textbf{7.36})}  \\
\cmidrule(lr){2-13}

\multirow{2}{*}{\texttt{prefix\_tuning}} & 61.62 & 86.98 & 91.06 & 92.46 & 87.07 & 71.12 & \textbf{95.18} & 90.13 & 66.13 & 78.16 & 91.46 & 82.85 \\
& \small{($\pm$4.93)} & \small{($\pm$18.91)} & \small{($\pm$4.09)} & \small{($\pm$9.55)} & \small{($\pm$15.58)} & \small{($\pm$6.06)} & \small{($\pm$0.54)} & \small{($\pm$29.23)} & \small{($\pm$\textbf{3.44})} & \small{($\pm$2.41)} & \small{($\pm$\textbf{2.44})} & \small{($\pm$8.79)}  \\
\cmidrule(lr){2-13}

\multirow{2}{*}{\texttt{lora}} & 63.99 & 87.59 & 92.60 & 93.11 & 88.48 & 80.26 & 94.99 & 90.72 & 70.63 &  82.46 & 91.85 & 85.15 \\
& \small{($\pm$20.64)} & \small{($\pm$4.29)} & \small{($\pm$4.39)} & \small{($\pm$3.77)} & \small{($\pm$2.57)} & \small{($\pm$9.28)} & \small{($\pm$8.48)} & \small{($\pm$19.31)} & \small{($\pm$8.65)} & \small{($\pm$8.86)} & \small{($\pm$21.68)} & \small{($\pm$10.17)} \\
\cmidrule(lr){2-13}

\multirow{2}{*}{\texttt{ia3}} & 63.03 & 86.19 & 92.32 & 91.88 & 86.41 & 76.89 & 94.42 & 90.65 & 66.85 & 78.52 & 91.56 & 83.52 \\
& \small{($\pm$21.39)} & \small{($\pm$5.08)} & \small{($\pm$3.94)} & \small{($\pm$3.73)} & \small{($\pm$13.46)} & \small{($\pm$7.17)} & \small{($\pm$2.13)} & \small{($\pm$29.16)} & \small{($\pm$9.69)} & \small{($\pm$10.11)} & \small{($\pm$21.94)} & \small{($\pm$11.62)} \\

\midrule

\textbf{Full Fine-tuning} & 63.66 & \textbf{87.63} & 90.20 & 92.81 & \textbf{91.92} & 78.77 & 94.81 & \textbf{91.20} & 68.87 & 82.91 & 91.33 & 84.91 \\

\bottomrule
\end{tabular}

    }
    \caption{Best performance ($\pm$ std. dev. across all hyper-parameters) of various supported single adapter methods (cf. §\ref{subsec:simple_peft}) applied to \texttt{roberta-base}, benchmarked against full fine-tuning. The best results and lowest std. dev. per task are highlighted in \textbf{bold}.} 

    \label{tab:roberta_results}
    \vspace{-5mm}
\end{table*}

\subsection{AdapterHub Ecosystem}

\libname is integrated into the extensive existing open-source ecosystem introduced by AdapterHub \citep{pfeiffer-etal-2020-adapterhub}.
Most prominently, this includes \href{https://AdapterHub.ml}{AdapterHub.ml} as a platform to share and discover pre-trained adapter modules.
\libname further broadens the possibilities for sharing adapters by integrating with HuggingFace's Model Hub, which has emerged as one of the primary platforms for open-sourcing model checkpoints.
The new Hub integration comes with programmatic methods of discovering and publishing pre-trained adapter modules, in addition to the previously available methods for downloading and saving adapters.

At the time of writing, users of \libname have access to over 700 pre-trained adapters. 

\section{Adapter Evaluation}

In addition to the ease of use aforementioned, we show that the adapter methods offered by our library are performant across a range of settings. To this end, we conduct evaluations on the single adapter implementations made available by \libname (see \S\ref{subsec:simple_peft}). We demonstrate the effectiveness of these methods against full fine-tuning on a variety of task types: extractive question answering~\cite{rajpurkar-etal-2018-know}, multiple choice classification~\cite{huang-etal-2019-cosmos}, sequence tagging~\cite{tjong-kim-sang-de-meulder-2003-introduction}, sequence to sequence summarization~\cite{narayan-etal-2018-dont}, sequence classification and sequence regression~\cite{wang-etal-2018-glue}. To make our evaluations reflective of user experience, we conduct them using the two most commonly used base model variants on AdapterHub: \textit{roberta-base}~\cite{liu2020roberta} and \textit{bart-base}~\cite{lewis-etal-2020-bart}. 

\rparagraph{Setup.}
We conduct a grid search over a range of common training hyper-parameters, varying learning rates between $\{10^{-5}, 10^{-4}, 5\cdot 10^{-4}, 10^{-4}, 10^{-3}\}$ and the number of epochs between $\{5, 10, 20, 30\}$. We also augment the grid with a number of adapter-specific hyper-parameters. These, along with the minimum and maximum trainable parameters added across the configurations, are detailed in \autoref{appendix:grid}. The highest attained performance (and the standard deviation of results across the grid) for the two chosen base models are outlined in Tables \ref{tab:roberta_results} and \ref{tab:bart_results}, respectively.

\rparagraph{Results.}
The obvious takeaway from our evaluations is that all adapter implementations offered by our framework are competitive with full model fine-tuning, across all task classes. Approaches that offer more tunable hyper-parameters (and thus allow for easy scaling) such as Bottleneck adapters, LoRA, and Prefix Tuning predictably have the highest topline performance, often surpassing full fine-tuning. However, extremely parameter-frugal methods like (IA)\textsuperscript{3}, which add $<0.005\%$ of the parameters of the base model, also perform commendably and only fall short by a small fraction. Finally, the Compacter is the least volatile among the single methods, obtaining the lowest standard deviation between runs on the majority of tasks.

\section{Conclusion}

We have presented \libname, a novel library for research and application of adapters.
Unlike comparable solutions, \libname equally focuses on the parameter-efficiency and modularity side of adapters.
Our library implements a diverse set of adapter methods under a unified interface which allows flexible configuration and mixing of different approaches.
We proposed a simple building block system for leveraging the modularity of adapters to build complex adapter setups.
\libname tightly integrates into the HuggingFace and AdapterHub ecosystems and its adapter implementations show performances competitive to full fine-tuning.

As research on adapters and LLMs continues to advance rapidly, our library will evolve as well.
Its extensibility makes \libname well prepared for the integration of new adapter methods and model architectures, both from us and the community.

\section*{Acknowledgements}
 This work has been funded by Huawei Technologies (Ireland) Co., Ltd. and German Research Foundation (DFG) as part of the Research Training Group KRITIS No. GRK 2222. We thank Martin Hentschel for his help building the demo web app for our library. We thank Francesco Piccinno and Srini Narayanan for their helpful comments and suggestions during the initial drafts of this paper. We also thank the many contributors and users of the \textit{AdapterHub} open-source framework. 

\bibliography{acl_anthology,custom}

\bibliographystyle{acl_natbib}

\appendix

\section{Description of Adapter Methods}\label{sec:adapter_descriptions}

\sparagraph{Bottleneck Adapters} introduce bottleneck modules in each layer of a Transformer model.
Generally, these adapter modules consist of a down-projection matrix $W_{down}$ that projects into a lower dimension $d_{bottleneck}$, a non-linearity $f$, an up-projection $W_{up}$ that projects back into the original hidden layer dimension and a residual connection $r$, i.e.:
$h \leftarrow W_{up} \cdot f(W_{down} \cdot h) + r$.

Depending on the specific configuration, these layers can be introduced at different locations and in \textbf{sequential} or \textbf{parallel} order relative to the adapted Transformer layer.
\libname provides pre-defined configurations for the sequential configurations of \citet{houlsbyParameterefficientTransferLearning2019} and \citet{pfeiffer-etal-2021-adapterfusion} as well as the parallel configuration of \citet{heUnifiedViewParameterEfficient2022}.



\rparagraph{Invertible Adapters} were proposed as part of MAD-X \citep{pfeiffer-etal-2020-mad} to learn language-specific transformations.
Embedding outputs are passed through an invertible adapter module in the forward direction before entering the first Transformer layer and in the inverse direction after leaving the last Transformer layer.

\rparagraph{Prompt Tuning \citep{lester-etal-2021-power}} is an approach to condition language models on a task-specific soft prompt.
While hard prompts consist of fixed textual descriptions prepended to the model's input \citep{brownLanguageModelsAre2020a}, these soft prompts are continuously optimized towards the target task via gradient descent.
Prompt tokens are prepended to the embedded input sequence.

\rparagraph{Prefix Tuning \citep{li-liang-2021-prefix}} is an extension of prompt tuning which prepends trainable prefix vectors $P^K$ and $P^V$ to the keys and values of the multi-head attention block inputs.
The prefix vectors have a configurable length and are not optimized directly but reparameterized via a bottleneck MLP in the built-in default configuration, following the original implementation.

\rparagraph{Compacter \citep{mahabadiCompacterEfficientLowRank2021a}} exchanges the linear down- and up-projection of a bottleneck adapter for a PHM layer\footnote{Parametrized hypercomplex multiplication layer.}. 
This PHM layer constructs its weight matrix from two smaller matrices, which reduces the number of parameters needed for the adapters.
These matrices can be factorized and shared between all adapter layers.
\libname provides pre-defined configurations for the Compacter and Compacter++ variants.

\rparagraph{LoRA \citep{huLoRALowRankAdaptation2022}} injects trainable low-rank decomposition matrices into the layers of a pre-trained model.
For any model layer expressed as a matrix multiplication of the form $h = W_0 x$, it performs a reparameterization such that: $h = W_0 x + \frac{\alpha}{r} B A x $.
Here, $A \in \mathbb{R}^{r\times k}$ and $B \in \mathbb{R}^{d\times r}$ are the decomposition matrices and $r$ is the low-dimensional rank of the decomposition.
With \libname, LoRA modules can be configured to be placed into the self-attention, intermediate, or output components of a Transformer layer.
Following \citet{huLoRALowRankAdaptation2022}, \libname provides a built-in method of merging LoRA modules with the original pre-trained weights of a model for inference without additional latency.

\rparagraph{(IA)³ \citep{liuFewShotParameterEfficientFineTuning2022}} introduces trainable vectors $l_W$ into different components of a Transformer model, which perform element-wise rescaling of inner model activations.
For any model layer expressed as a matrix multiplication of the form $h = W x$, it therefore performs an element-wise multiplication with $l_W$, such that: $h = l_W \odot W x$.

\begin{table*}[ht!]
    \centering
    \def\arraystretch{0.87}
    \resizebox{0.76\linewidth}{!}{
    \begin{tabular}{rcccccccccccc}

\toprule
\multirow{3}{*}{\textbf{Method}} & \multicolumn{5}{c}{\textbf{Sequence Classification}} & \textbf{Regression} &      \textbf{Extract. QA} & \textbf{Seq2Seq} & \multirow{3}{*}{\textbf{Avg.}} \\
\cmidrule(l){2-6} \cmidrule(l){7-7} \cmidrule(l){8-8} \cmidrule(l){9-9} 
 &  \textbf{CoLA} &  \textbf{MRPC} &  \textbf{QNLI} &  \textbf{RTE} &  \textbf{SST-2} &  \textbf{STS-B} & \textbf{SQuAD v2} & \textbf{XSum} & \\

& \texttt{Dev MCC} & \texttt{Dev F1} & \texttt{Dev Acc.} & \texttt{Dev Acc.} & \texttt{Dev Acc.} & \texttt{Dev PCC} & \texttt{Dev F1} & \texttt{Rouge 2} & \\
 
\midrule

\multirow{2}{*}{\texttt{double\_seq\_bn}} & 61.80  & 91.62 & 92.18 & 73.65 & 94.04 & 89.89 & 79.37 &  17.49  &  75.00\\
& \small{($\pm 21.47$)} & \small{($\pm$3.73)} & \small{($\pm$15.35)} & \small{($\pm$8.26)} & \small{($\pm$13.89)} & \small{($\pm$23.18)} & \small{($\pm$12.17)} & \small{($\pm$1.51)} & \small{($\pm$10.86)} \\
\cmidrule(lr){2-10}

\multirow{2}{*}{\texttt{seq\_bn}} &53.29  & 91.16 & 92.11 & 75.45 & 93.46 & 89.44 & 79.47 & 17.76 & 74.02  \\
& \small{($\pm 18.82$)} & \small{($\pm$3.59)} & \small{($\pm$6.29)} & \small{($\pm$6.84)} & \small{($\pm$\textbf{1.09})} & \small{($\pm$20.38)} & \small{($\pm$10.41)} & \small{($\pm$1.60)} & \small{($\pm$7.55)} \\
\cmidrule(lr){2-10}

\multirow{2}{*}{\texttt{par\_bn}} & 55.58 & 90.75 & 92.12 & 75.09 & 94.08 & 89.58 & 79.63 & \textbf{18.26}  & 74.39  \\
& \small{($\pm$19.70)} & \small{($\pm$3.46)} & \small{($\pm$14.42)} & \small{($\pm$7.91)} & \small{($\pm$14.39)} & \small{($\pm$\textbf{10.86})} & \small{($\pm$\textbf{8.19})} & \small{($\pm$ 1.19)} & \small{($\pm$10.01)} \\
\cmidrule(lr){2-10}

\multirow{2}{*}{\texttt{compacter}} & 44.81 & 87.44 & 91.45 & 69.68 & 93.35 & 87.62 & 74.33 & 13.55 & 70.28 \\
& \small{($\pm$12.06)} & \small{($\pm$4.30)} & \small{($\pm$\textbf{3.14})} & \small{($\pm$\textbf{3.85})} & \small{($\pm$1.18)} & \small{($\pm$14.09)} & \small{($\pm$9.81)} & \small{($\pm$5.29)} & \small{($\pm$\textbf{6.72})} \\
\cmidrule(lr){2-10}

\multirow{2}{*}{\texttt{prefix\_tuning}} & 48.57 & \textbf{92.31} & 91.61 & 75.95 & 93.58 & 90.27 & 80.45 & 16.81 & 73.69 \\
& \small{($\pm$\textbf{8.23})} & \small{($\pm$3.65)} & \small{($\pm$6.97)} & \small{($\pm$5.63)} & \small{($\pm$1.92)} & \small{($\pm$13.72)} & \small{($\pm$8.73)} & \small{($\pm$4.94)} & \small{($\pm$6.73)} \\
\cmidrule(lr){2-10}

\multirow{2}{*}{\texttt{lora}} & 55.94 & 91.61 & 92.53 & 76.05 & 94.03 & 89.57 & 78.81 & 17.21 & 74.47 \\
& \small{($\pm$21.93)} & \small{($\pm$4.17)} & \small{($\pm$4.31)} & \small{($\pm$8.67)} & \small{($\pm$1.31)} & \small{($\pm$33.03)} & \small{($\pm$10.47)} & \small{($\pm$\textbf{1.13})} & \small{($\pm$10.63)} \\
\cmidrule(lr){2-10}

\multirow{2}{*}{\texttt{ia3}} & 50.65 & 89.68 & 90.80 & 72.92 & 94.22 & 89.05 & 72.46 & 13.51 & 71.66 \\
& \small{($\pm$13.64)} & \small{($\pm$\textbf{2.79})} & \small{($\pm$7.29)} & \small{($\pm$5.83)} & \small{($\pm$6.04)} & \small{($\pm$21.27)} & \small{($\pm$8.46)} & \small{($\pm$4.08)} & \small{($\pm$8.68)} \\

\midrule

\textbf{Full Fine-tuning} & \textbf{62.80}  & 90.4 & \textbf{94.9} & \textbf{87.0} & \textbf{96.6} & \textbf{91.2} & \textbf{89.2} & 17.73  & \textbf{86.84} \\

\bottomrule
\end{tabular}

    }
    \caption{Best performance ($\pm$ std. dev. across all hyper-parameters) of various supported single adapter methods (refer §\ref{subsec:simple_peft}) applied to \texttt{facebook/bart-base}, benchmarked against full fine-tuning. The best results and lowest std. dev. per task are highlighted in \textbf{bold}.} 

    \label{tab:bart_results}
\end{table*}

\label{appendix:grid}

\begin{table}[h!]
    \centering
    \def\arraystretch{0.87}
    \resizebox{\linewidth}{!}{
    \begin{tabular}{rccll}

\toprule
\multirow{2}{*}{\textbf{PEFT Method}} 
 &  \textbf{Attribute Name} &  \textbf{Range} &  \multicolumn{2}{c}{\textbf{Added Params.}} \\

&  &  & \texttt{Min} & \texttt{Max} \\
 
\midrule

\texttt{double\_seq\_bn} & \texttt{reduction\_factor} & $\{2, 16, 64\}$ & \texttt{461,088} & \texttt{\textbf{14,183,424}} \\

\cmidrule(lr){2-5}

\texttt{seq\_bn} & \texttt{reduction\_factor} & $\{2, 16, 64\}$ & \texttt{230,544} & \texttt{7,091,712} \\   
\cmidrule(lr){2-5}

\texttt{par\_bn} & \texttt{reduction\_factor} & $\{2, 16, 64\}$ & \texttt{230,544 }
 & \texttt{7,091,712} \\
\cmidrule(lr){2-5}

\multirow{2}{*}{\texttt{compacter}} & \texttt{reduction\_factor} & $\{4, 16\}$ & \multirow{2}{*}{\texttt{58,816}} & \multirow{2}{*}{\texttt{69,184}} \\
& \texttt{phm\_dim} & $\{4, 8\}$ &  &  \\
\cmidrule(lr){2-5}

\multirow{2}{*}{\texttt{prefix\_tuning}} & \texttt{bottleneck\_size} & $\{32, 128, 512\}$ & \multirow{2}{*}{\texttt{636,704}} & \multirow{2}{*}{\texttt{10,002,944}} \\
& \texttt{prefix\_length} & $\{5, 50, 200\}$ &  &  \\
\cmidrule(lr){2-5}

\texttt{lora} & \texttt{r} & $\{4, 8, 16, 64, 200\}$ & \texttt{147,456} & \texttt{7,372,800} \\
\cmidrule(lr){2-5}

\texttt{ia3} & $-$ & $-$ & \texttt{\textbf{55,296}} & \texttt{55,296} \\

\bottomrule
\end{tabular}

    }
    \caption{Adapter-specific attributes (name as in framework) and their values that we used for grid-search along with the minimum and maximum possible parameters added over the resultant grid. Extreme values are highlighted in \textbf{bold}.}
    \label{tab:search_space}
\end{table}

\newpage

\section{Adapter Evaluation Results}

\begin{figure}
    \centering
    \includegraphics[width=\linewidth]{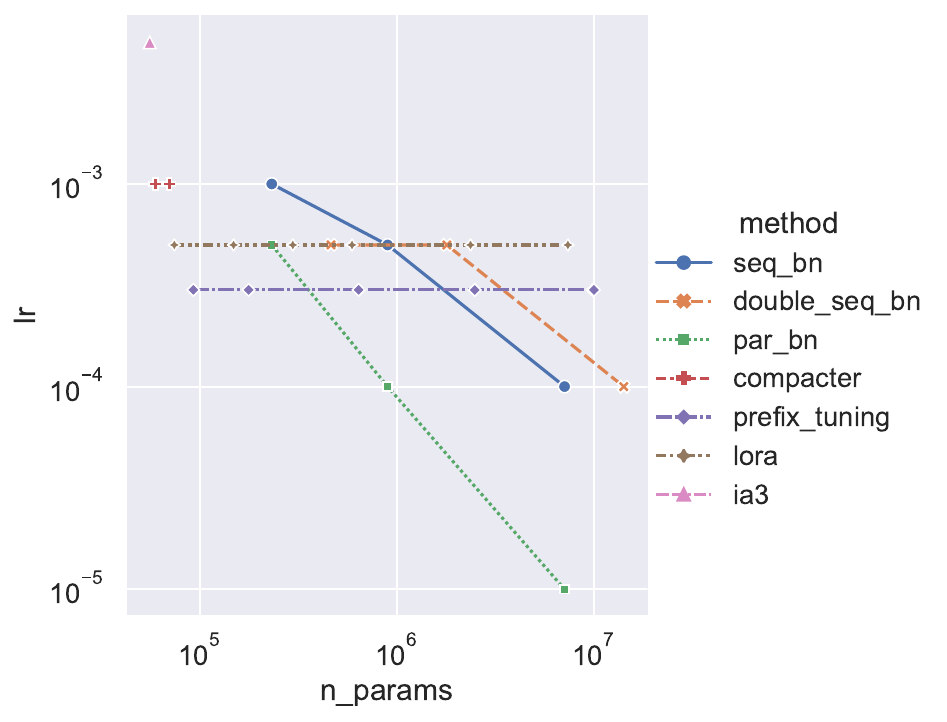}
    \caption{Best performing learning rate--capacity combinations for evaluated single adapter methods. Lower capacity adapters perform better with higher learning rates.}
    \label{fig:nparams_vs_lr}
\end{figure}

Best task performance for all task--adapter method combinations are presented in \cref{tab:roberta_results} with RoBERTa as base and model and in \cref{tab:bart_results} with BART as base model. All adapter methods provided by \libname are competitive to full fine-tuning.
\Cref{tab:search_space} gives an overview of the evaluated adapter configurations.
\cref{fig:nparams_vs_lr} plots best performing learning rate--capacity combinations for evaluated single adapter methods. Lower capacity adapters perform better with higher learning rates.

\begin{figure*}[!t]
    \centering
    \includegraphics[width=\textwidth]{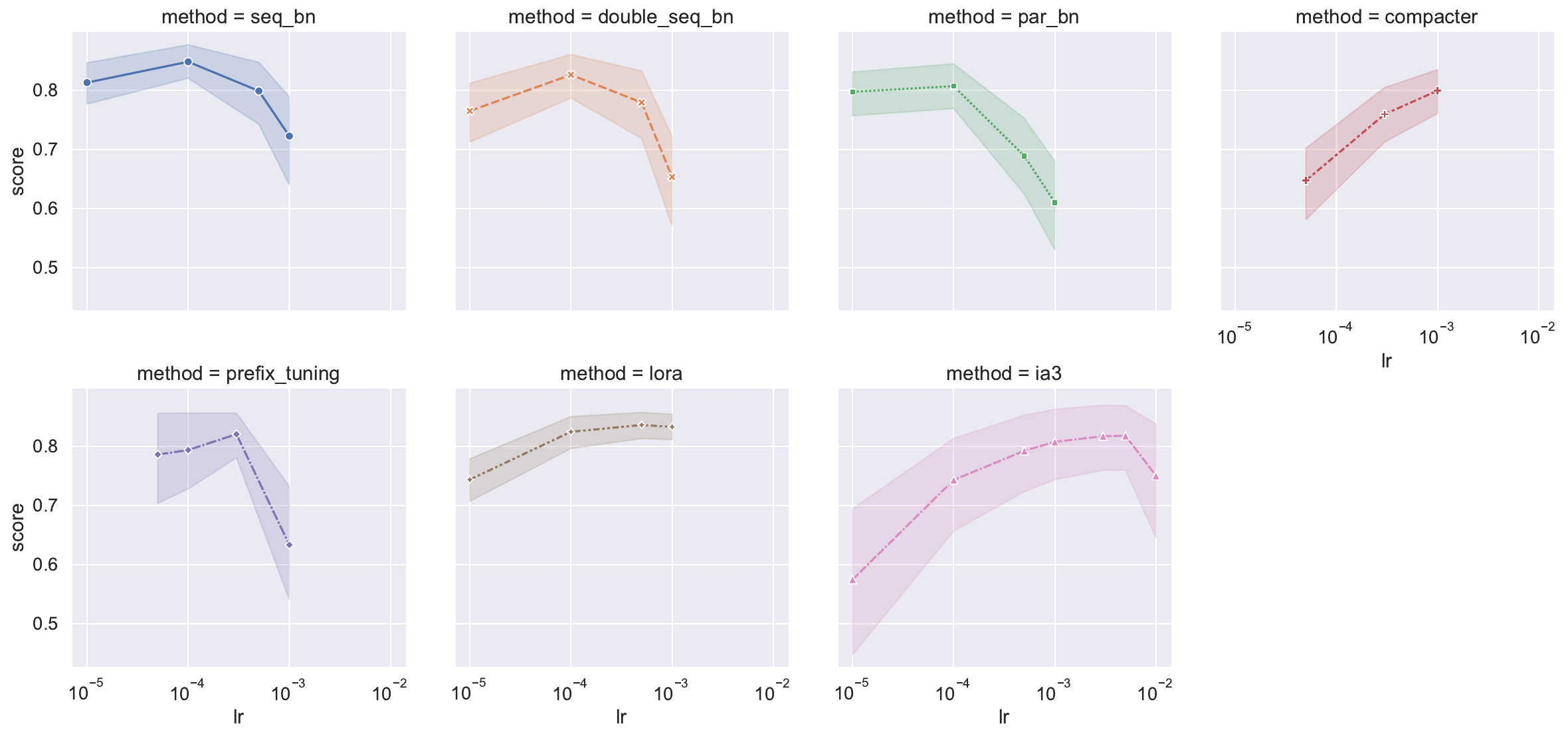}
    \caption{Mean task performance by learning rate for evaluated single adapter methods.}
    \label{fig:ls_vs_score}
\end{figure*}

\cref{fig:ls_vs_score} plots mean task performance by learning rate for evaluated single adapter methods. For bottleneck adapters, the best-performing learning rate is $10^{-4}$. Compacter performs best with a learning rate of $10^{-3}$, prefix tuning with $3 \cdot 10^{-4}$, LoRA with $5 \cdot 10^{-4}$, and (IA)³ with $5 \cdot 10^{-3}$. These results align with the learning rates proposed by the respective adapter method authors. 

\end{document}